\documentclass[letterpaper, 10 pt, conference]{ieeeconf}  

\IEEEoverridecommandlockouts                              

\usepackage{graphics} 
\usepackage[pdftex]{graphicx}
\usepackage{circuitikz}
\usepackage{tikz}
\usepackage{tikzscale}
\usetikzlibrary{arrows}
\usepackage{scalefnt}
\graphicspath{ {./images/} }
\usepackage{amsmath}
\usepackage{float}
\usepackage{circuitikz}
\usepackage{pgfplots}
\usepgfplotslibrary{units}
\usepackage{filecontents}
\usepackage{tikz,pgfplots,filecontents,amsmath}
\usepackage{algorithmic}
\usepackage{amssymb}    

\usepackage{bm,times}
\usepackage{bm} 
\usepackage{color} 
\usepackage{hyperref}
\usepackage{cancel}
\usepackage[export]{adjustbox}
\usepackage{mathtools}
\usepackage{subfig}
\usepackage{fixltx2e}
\usepackage[capitalise]{cleveref}
\usetikzlibrary{calc}
\newcommand{\tikzmark}[1]{\tikz[overlay,remember picture] \node (#1) {};}
\newcommand{\DrawBox}[4][]{%
    \tikz[overlay,remember picture]{%
        \coordinate (TopLeft)     at ($(#2)+(-0.9em,0.9em)$);
        \coordinate (BottomRight) at ($(#3)+(0.6em,-0.45em)$);
        \path (TopLeft); \pgfgetlastxy{\XCoord}{\IgnoreCoord};
        \path (BottomRight); \pgfgetlastxy{\IgnoreCoord}{\YCoord};
        \coordinate (LabelPoint) at ($(\XCoord+4.5em,\YCoord+5.3em)!0.5!(TopLeft)$);
        \draw [red,#1] (TopLeft) rectangle (BottomRight);
        \node [below, #1, fill=none, fill opacity=1] at (LabelPoint) {#4};
    }
}

\newcommand{\DrawBoxx}[4][]{%
    \tikz[overlay,remember picture]{%
        \coordinate (TopLeft)     at ($(#2)+(-0.9em,0.9em)$);
        \coordinate (BottomRight) at ($(#3)+(0.45em,-0.45em)$);
        \path (TopLeft); \pgfgetlastxy{\XCoord}{\IgnoreCoord};
        \path (BottomRight); \pgfgetlastxy{\IgnoreCoord}{\YCoord};
        \coordinate (LabelPoint) at ($(\XCoord-7.8em,\YCoord)!0.5!(BottomRight)$);
        \draw [red,#1] (TopLeft) rectangle (BottomRight);
        \node [below, #1, fill=none, fill opacity=1] at (LabelPoint) {#4};
    }
}

\usepackage{pdfpages}
\usepackage{graphicx}
\usepackage{caption}
\usepackage{booktabs} 
\usepackage{amsmath,bm,mathtools,upgreek}
\usepackage{amssymb,stmaryrd}
\usepackage{pgfplots}
\usepackage{prettyref}
\newrefformat{fig}{Figure~[\ref{#1}]}
\newsavebox\IBoxA \newsavebox\IBoxB \newlength\IHeight
\newcommand\TwoFig[6]{
  \sbox\IBoxA{\includegraphics[width=0.3\textwidth]{#1}}
  \sbox\IBoxB{\includegraphics[width=0.3\textwidth]{#4}}%
  \ifdim\ht\IBoxA>\ht\IBoxB
    \setlength\IHeight{\ht\IBoxB}%
  \else\setlength\IHeight{\ht\IBoxA}\fi
  \begin{figure}[!htb]
  \minipage[t]{0.3\textwidth}\centering
  \includegraphics[height=\IHeight]{#1}
  \caption{#2}\label{#3}
  \endminipage\hfill
  \minipage[t]{0.3\textwidth}\centering
  \includegraphics[height=\IHeight]{#4}
  \caption{#5}\label{#6}
  \endminipage 
  \end{figure}%
}
\makeatletter 
\newcommand*{\b@xplus}[1][+]{\ooalign{%
    $\m@th\vcenter{\hbox{$\m@th#1$}}$\cr%
    \hidewidth$\m@th\boxempty$\hidewidth\cr}} 
\renewcommand*{\boxplus}{\mathbin{\b@xplus}} 
\renewcommand*{\boxminus}{\mathbin{\b@xplus[-]}} 
\makeatother
\renewcommand{\vec}[1]{\bm{\mathrm{#1}}}

\newtheorem{theorem}{Theorem}[section]

\newtheorem{definition}{Definition}[section]
\title{\LARGE \bf
Passivity-based control for haptic teleoperation of a legged manipulator in presence of time-delays
}

\author{Mattia Risiglione$^{{1},{2}}$, Jean-Pierre Sleiman$^{1}$, Maria Vittoria Minniti$^{1}$,\\ Burak Çizmeci$^{1}$, Douwe Dresscher$^{3}$ and Marco Hutter$^{1}$
\thanks{This research was supported in part by the Swiss National Science Foundation through the National Centre of Competence in Research Robotics (NCCR Robotics), in part by the Swiss National Science Foundation through the National Centre of Competence in Digital Fabrication (NCCR dfab), in part by TenneT, and in part by Armasuisse Science and Technology.}
\thanks{$^{1}$Authors are with Robotics System Lab (RSL),  ETH Zurich, Switzerland.} %
\thanks{$^{2}$Author is with Dynamic Legged Systems Lab, Istituto Italiano di Tecnologia (IIT). \tt\small mattia.risiglione@iit.it.} %
\thanks{$^{3}$Author is with Faculty of Electrical Engineering, Mathematics and Computer Science, University of Twente, The Netherlands.}%
}

\begin{document}

\maketitle
\thispagestyle{empty}
\pagestyle{empty}


\begin{abstract}
When dealing with the haptic teleoperation of multi-limbed mobile manipulators, the problem of mitigating the destabilizing effects arising from the communication link between the haptic device and the remote robot has not been properly addressed. In this work, we propose a passive control architecture to haptically teleoperate a legged mobile manipulator, while remaining stable in the presence of time delays and frequency mismatches in the master and slave controllers. At the master side, a discrete-time energy modulation of the control input is proposed. At the slave side, passivity constraints are included in an optimization-based whole-body controller to satisfy the energy limitations. A hybrid teleoperation scheme allows the human operator to remotely operate the robot's end-effector while in stance mode, and its base velocity in locomotion mode. The resulting control architecture is demonstrated on a quadrupedal robot with an artificial delay added to the network. 
\end{abstract}


\section{INTRODUCTION}
Teleoperation offers solutions to extend the sensing and manipulation capabilities of human operators performing long-distance tasks. Recent advances of robotic platforms with loco-manipulation skills represent a promising resource for enlarging the operational workspace at the remote site. In particular, legged robots have the ability to locomote over rough terrains, interact with objects and perceive their surroundings. Hence, these platforms can be used as avatar systems to enhance the human's full immersion in the remote environment and to safely perform complex operations in hazardous scenarios.

Among all the feedback information that stimulates the human senses, haptics has been shown to be of paramount importance \cite{wildenbeest2012impact}. When signals flow bilaterally between the device handled by the human operator, named master device, and the remote robot being teleoperated, named slave, we refer to bilateral teleoperation.
In such an application, computing control laws with delayed and/or incomplete information can easily destabilize the system. 
Besides latency and packet losses, other components can negatively influence the stability of the system, such as the grasping force exerted by the user, haptic feedback gains and environment stiffness. 
Closed-loop stability in haptics has been investigated since the late 80s\cite{anderson1988bilateral}. In this context, passivity theory has been applied, with many proposed solutions differing in the way energy flows are monitored and limited \cite{hokayem2006bilateral}, \cite{muradore2016review}. Among these, the Two-Layer architecture from \cite{franken2011bilateral} uses virtual storage elements to observe the energy exchanged between the master and slave device. Such an approach is made of two stages: first, the control law is computed, next energy dissipation is applied to act against a passivity loss of the bilateral controller. With respect to other methods such as wave variables \cite{niemeyer1991stable} or time-domain approaches \cite{ryu2004stable}, passive controllers based on the Two-Layer approach (i.e., \cite{franken2011bilateral}, \cite{sartoriTSM19}, \cite{franco2018transparency}) do not require  any encoding of the power variables, nor rely on any assumptions on the sampling rate. Additionally, the transparency of the system can be improved by optimizing a transparency metric in the passivity layer \cite{franco2018transparency}.

\begin{figure}[t]
\centering
\includegraphics[scale=0.18]{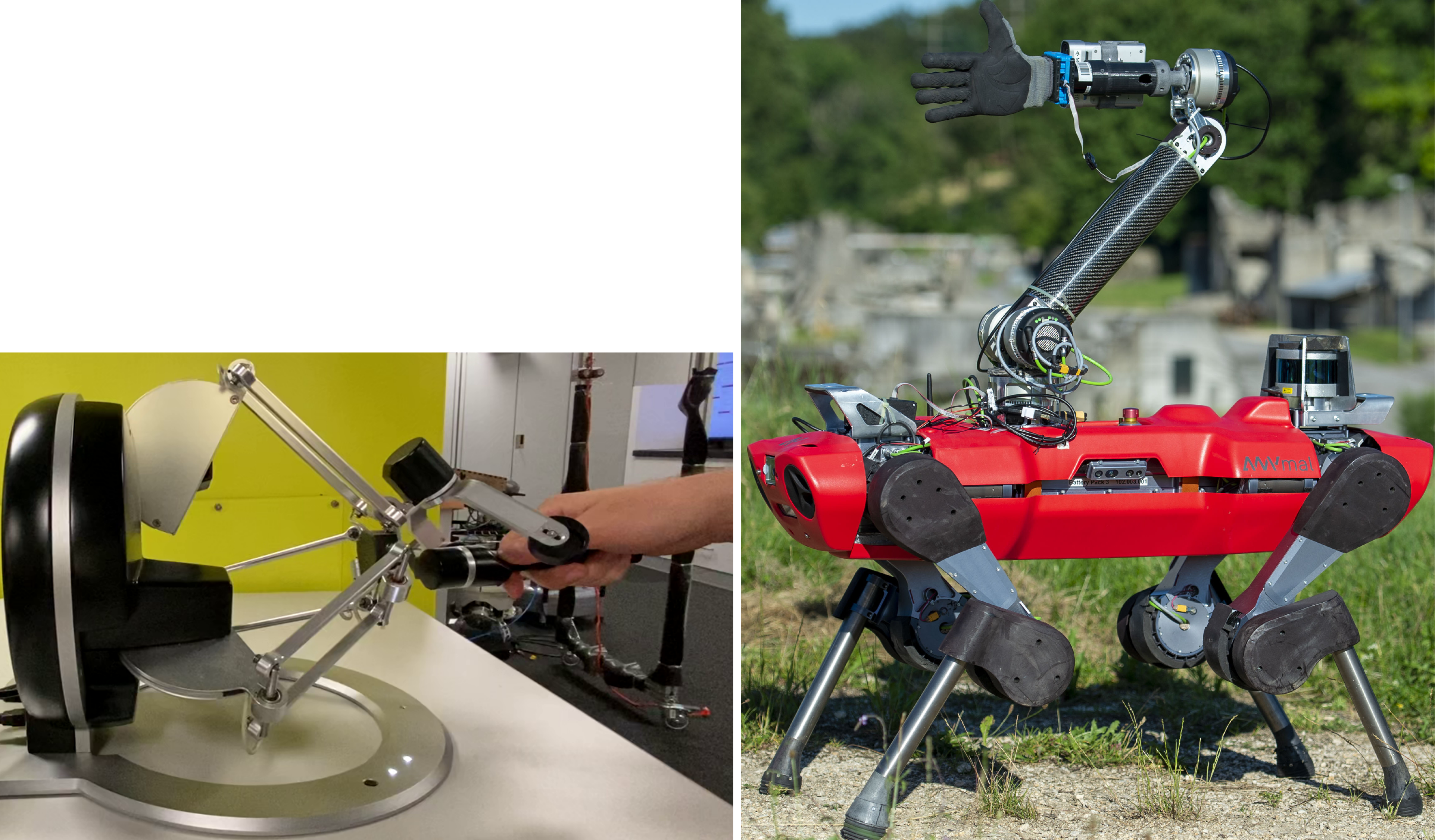}
\caption{Experimental setup with Omega6 as haptic device at the master side \textit{(left)} and the ANYmal  C  platform  equipped  with  a robotic arm at the slave side \textit{(right)}.}
\vspace{-0.6cm}
\end{figure}

For multi-degree of freedom (DOF) floating-base systems, such as quadrupeds or humanoids, optimization-based techniques are prominently adopted for whole-body control (WBC) design \cite{sentis2005synthesis}, \cite{bellicoso2016perception}. Such control schemes optimize control objectives for multiple tasks, while handling physical constraints, such as actuation limits, friction cone constraints, etc. Passive whole-body controllers have been proposed to compliantly and safely interact with the environment \cite{fahmi2019passive}, \cite{henze2016passivity}. However, in a haptic teleoperation scenario, existing work has focused on bounding \cite{xin2020bounded} and mapping \cite{gomesRPMMI19} the operator commands to ensure safe teleoperation and maintain balance, rather than designing a controller that is inherently robust against the effects of time-delays. One way of addressing this problem is through an energy dissipation performed at the output of the whole-body controller, however this could interfere with the constrained optimal solution and render it infeasible.
Here, instead of having a control computing layer (\textit{transparency layer}) and a dissipation layer (\textit{passivity layer}), we directly create an optimal passivity controller that respects the energy constraints. 

Hence, we propose a control framework for the time-delayed haptic teleoperation of a legged robot, that is in particular capable of mitigating the destabilizing effects caused by time delays in the communication link. To this end, we provide new insights for the formulation of energy constraints in optimization-based whole-body controllers. We then validate the effectiveness of our approach with a set of hardware experiments performed on a quadrupedal mobile manipulator, where the robot is teleoperated during end-effector or base control while introducing artificial yet realistic delays in the network. It is worth noting that the resulting architecture is not specific for quadrupedal robots, but is general enough to be applied to any floating-base system, such as humanoids or wheeled robots.


\section{BILATERAL TELEOPERATION PROBLEM}
\label{sec:bilateral_teleoperator_setup}
In this section, we first provide some necessary background on passivity theory. Afterwards, we describe the considered bilateral teleoperator, and how the passivity requirements can be satisfied with the use of energy tanks.
\subsection{Background on passivity theory}
Consider a state-space system given by:
\begin{equation}
\begin{split}
\vec{\dot{x}} &= \vec{f}(\vec{x},\vec{u})\\
\vec{y}       &= \vec{h}(\vec{x},\vec{u}),\\
\end{split}
\label{generic state space system}
\end{equation}
where $\vec{x} \in \mathcal{X} \subseteq \mathbb{R}^{n_x}$,
$\vec{u} \in \mathcal{U} \subseteq \mathbb{R}^{n_u}$,
$\vec{y} \in \mathcal{Y} \subseteq \mathbb{R}^{n_y}$.
\begin{definition}[\cite{tadmor05}]
\label{definition system passivity}
System \eqref{generic state space system} is said to be passive with respect to the input $\vec{u}$ and output $\vec{y}$ if there exists a continuous function, $\mathcal{H}(\cdot) \ge 0$, called storage function, such that
\begin{equation}
\mathcal{H}(t) - \mathcal{H}(0) \le \int_{0}^{t} \vec{u}^T(s) \vec{y}(s)\; ds  ,
\label{definition passivity}
\end{equation}
 $\forall \mbox{ }t \ge 0$, for all input signals $u \in \mathcal{U} \subseteq \mathbb{R}^{n_u}$, for all initial states $x(0) \in \mathcal{X}$ and $n_u=n_y$. \end{definition}
 Equivalently, we can define system \eqref{generic state space system} to be passive if $\exists \mathcal{H}(\cdot) \ge 0$ such that
 $
\dot{\mathcal{H}}(t) \le \vec{u}^T(t)\vec{y}(t)  
$.
\begin{theorem}[\cite{ebenbauer2009dissipation}]
\label{feedback interconnection}
Consider the two systems
\begin{equation}
\Sigma_i :
\begin{cases}
\vec{\dot{x}}_i = \vec{f}_i(\vec{x}_i,\vec{u}_i)\\
\vec{y}_i       = \vec{h}_i(\vec{x}_i,\vec{u}_i),    
\end{cases}
\end{equation}
where $i = 1,2$, $n_u = n_y$, interconnected with the following power-preserving feedback interconnection:
\begin{equation}
\begin{bmatrix}
\vec{u}_1 \\
\vec{u}_2 \\
\end{bmatrix} = 
\begin{bmatrix}
\vec{0}_{n_u\times n_u} & \mathcal{\vec{W}}\\
-\mathcal{\vec{W}} & \vec{0}_{n_u\times n_u}
\end{bmatrix} 
\begin{bmatrix}
\vec{y}_1 \\
\vec{y}_2,\\
\end{bmatrix}
\label{power preserving interconnection prototype}
\end{equation}
where $\mathcal{\vec{W}} := diag(w_1,..., w_{{n}_u})$, $\mathcal{\vec{W}} \succcurlyeq 0$. If $\Sigma_i$ is passive with respect to the mapping $\vec{u}_i \to \vec{y}_i$ with storage function $\mathcal{H}_{\Sigma_i}$ for $i=1,2$, then the interconnected system $\Sigma_1 \wedge \Sigma_2$ is also passive and the overall storage function is such that
\begin{equation}
\begin{cases}
\mathcal{H}_{\Sigma_1 \wedge \Sigma_2} =  \mathcal{H}_{\Sigma_1} + \mathcal{H}_{\Sigma_2}\\
\dot{\mathcal{H}}_{\Sigma_1 \wedge \Sigma_2} = 0.
\end{cases}
\end{equation}
\end{theorem}

\subsection{Control by interconnection}
\label{subsec:control_by_interconnection}
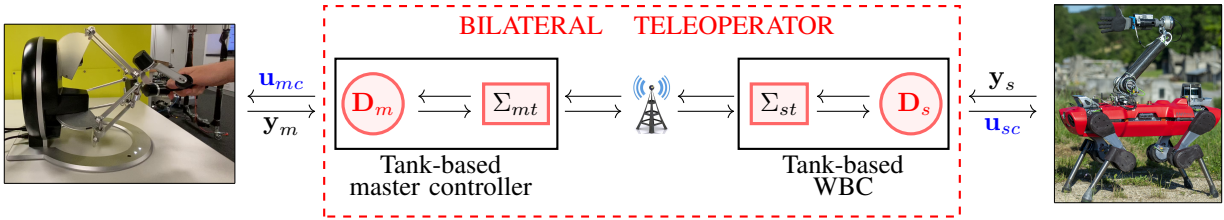
\begin{figure*}[t]
\centering
\begin{tikzpicture}[]
\node[inner sep=0pt] (russell) at (-7.97,-0.7)
  {\includegraphics[frame,scale=.08]{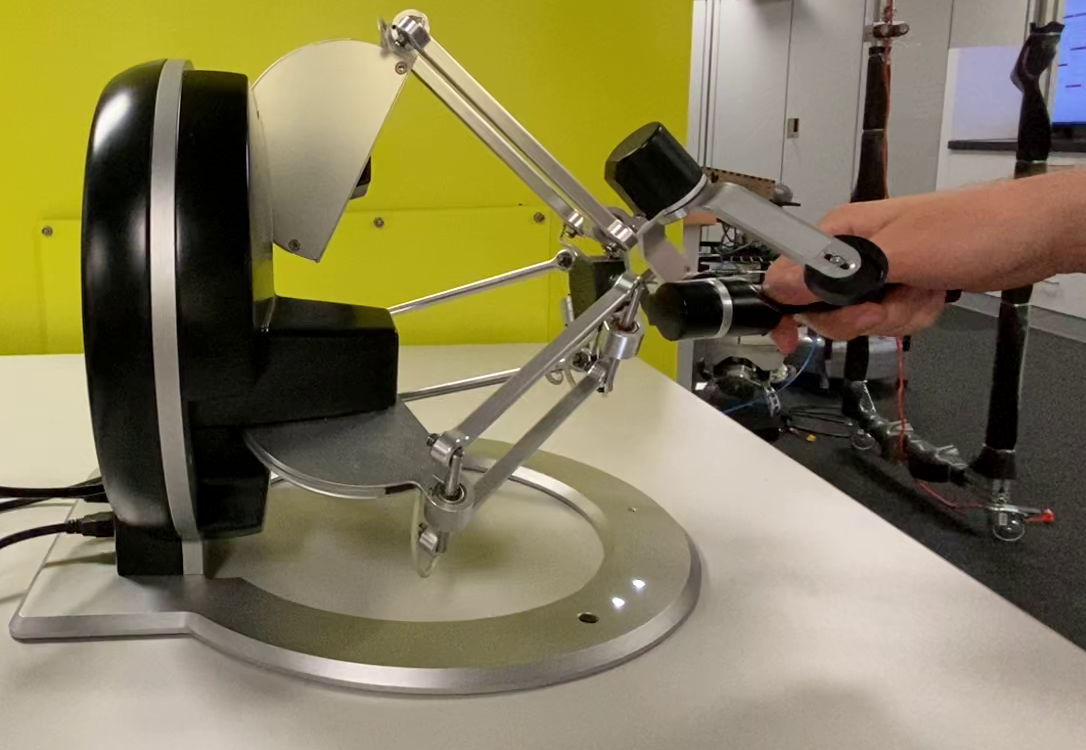}};    
\node(b)[rectangle, draw=red!60, fill=red!5, very thick, minimum size=5mm, right of=a, minimum width=0.5cm,minimum height=0.5cm] at (-3.7, -0.7) {$\Sigma_{mt}$};
\filldraw[color=red!60, fill=red!5, very thick](-4.6,-.7) circle (0.4);
\node(c)[rectangle, draw=red!60, fill=red!5, very thick, minimum size=5mm, right of=a, minimum width=0.5cm,minimum height=0.5cm] at (-0.2, -0.7) {$\Sigma_{st}$};
\filldraw[color=red!60, fill=red!5, very thick](2.55,-0.7) circle (0.4);
\draw[red,thick,dashed] (-5.25,-2.2) rectangle (3.2,0.6);
\draw[black,thick] (-5.1,-1.3) rectangle (-2.15,-0.1);
\draw[black,thick] (0.26,-1.3) rectangle (3.05,-0.1);
\node[text=black](d) at (-3.7,-1.5) {Tank-based};
\node[text=black](d) at (-3.7,-1.8) {master controller};
\node[text=black](d) at (1.65,-1.5) {Tank-based};
\node[text=black](d) at (1.65,-1.8) {WBC};

\draw[<-] (-6.27,-0.6) -- (-5.37,-0.6);
\draw[->] (-6.27,-0.8) -- (-5.37,-0.8);
\draw[<-] (-4.0,-0.6) -- (-3.3,-0.6);

\draw[->] (-4.0,-0.8) -- (-3.3,-0.8);
\draw[<-] (1.3,-0.6) -- (2.0,-0.6);
\draw[->] (1.3,-0.8) -- (2.0,-0.8);
\draw[<-] (-0.55,-0.6) -- (0.2,-0.6);
\draw[->] (-0.55,-0.8) -- (0.2,-0.8);
\node[text=red](d) at (-4.6,-0.7) {$\vec{D}_{m}$};
\node[text=red](d) at (2.6,-0.7) {$\vec{D}_{s}$};
\draw[<-] (-2.05,-0.6) -- (-1.3,-0.6);
\draw[->] (-2.05,-0.8) -- (-1.3,-0.8);

\draw[->] (4.22,-0.6) -- (3.32,-0.6);
\draw[->] (3.32,-0.8) -- (4.22,-0.8);

\node(d) at (-5.82,-1) {$\vec{y}_{m}$};
\node[text=blue](c) at (-5.82,-0.4) {$\vec{u}_{mc}$};
\node[text=blue](f) at (3.77,-1) {$\vec{u}_{sc}$};
\node(g) at (3.77,-0.4) {$\vec{y}_s$};
\node[inner sep=0pt] (russell) at (5.57,-0.7)
    {\includegraphics[frame,scale=.115]{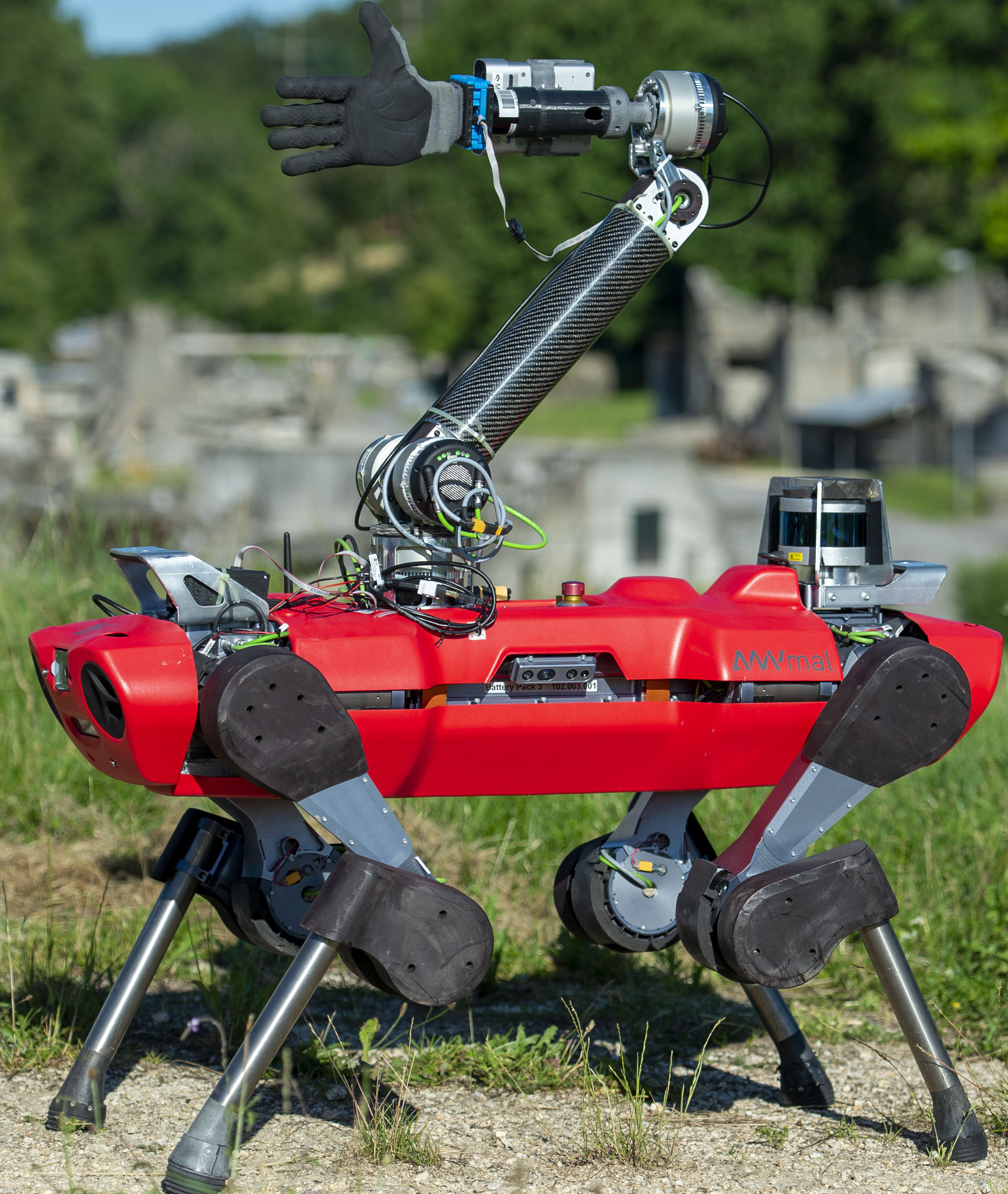}};
\node[inner sep=0pt] (russell) at (-0.9,-0.7)
    {\includegraphics[scale=.05]{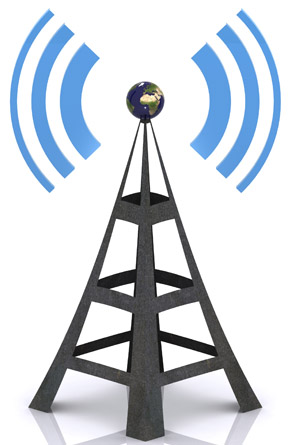}};  
\node[text=red, thick](f) at (-2.5,0.35) {BILATERAL};
\node[text=red, thick](f) at (0.2,0.35) {TELEOPERATOR};

\end{tikzpicture}
\caption{Diagram of main elements interacting, from left to right: the human operator and the haptic device, the bilateral teleoperator, the slave and the environment.} 
\label{main blocks}
\end{figure*}

The closed-loop system in \cref{main blocks} is passive if its interacting components satisfy \eqref{definition passivity}, and are interconnected according to \eqref{power preserving interconnection prototype}.
Being mechanical systems, the master and slave are passive, as it can be deduced considering the system Hamiltonian as a storage function.
Therefore, a passive control unit needs to be designed to convey energy between the haptic interface and the remote robot. 
This can be done by ensuring that the assumptions of Theorem \eqref{feedback interconnection} are satisfied, i.e. that the passivity requirement \eqref{definition passivity}
holds for the bilateral teleoperator and this results passively interconnected to the master and slave devices.
In the next section, we show how to preserve the passivity requirements using the concept of virtual energy tanks \cite{tadele2014combining}.

\subsection{Passivity control with energy tanks}
\label{subsec:passivity_control_with_energy_tanks}
Energy tanks are dynamical systems that are connected to the master and slave systems and exchange energy between each other.  The energy stored inside each tank is equal to the one transferred from the other tank, and exchanged with the physical world, and represents the energy budget that can be used to generate desired active behaviours. We denote these two storage elements as: $\Sigma_{mt}(u_{mt}, y_{mt})$ and $\Sigma_{st}(u_{st}, y_{st})$.
The energy tanks are modeled as dynamical systems with the following dynamics:
\begin{subequations}
\begin{align}
\Sigma_{mt} &:
\begin{cases}
\dot{x}_{mt} = u_{mt} + \left(\frac{P_{mt}^{+} - P_{mt}^{-}}{x_{mt}} \label{master tank}\right)\\
y_{mt} = \partial_{x_{mt}}{\mathcal{H}}_{mt} = x_{mt}
\end{cases}    \\
\Sigma_{st} &:
\begin{cases}
\dot{x}_{st} = u_{st} + \left(\frac{P_{st}^{+} - P_{st}^{-}}{x_{st}}\right)\\
y_{st} = \partial_{x_{st}}{\mathcal{H}}_{st} = x_{st}, \label{slave tank}
\end{cases}
\end{align}    
\end{subequations}
where $\mathcal{H}_{mt} = \frac{1}{2} {x_{mt}}^2$, $\mathcal{H}_{st} = \frac{1}{2} {x_{st}}^2$ are the tanks storage functions, and $P^+ \ge 0$ and $P^-\ge 0$ indicate the incoming and outgoing power exchanged between the two tanks. In this work, energy is exchanged only if present inside the tank, according to the energy transfer protocol described in \cite{franken2011bilateral}.
The systems $\Sigma_{mt}$ and $\Sigma_{st}$ are connected to the master and slave ports, respectively, as follows:

\begin{equation}
\begin{bmatrix}
\vec{u}_{mc} \\
u_{mt} \\
\vec{u}_{sc} \\
u_{st} 
\end{bmatrix} = 
\begin{bmatrix}
\tikzmark{left1}\vec{0} & \frac{\vec{w}_m}{x_{mt}} & \vec{0} & 0\\
-\frac{\vec{w}_m^T}{x_{mt}} & 0\tikzmark{right1}  & \vec{0} & 0\\
\vec{0} & 0 & \tikzmark{left2}\vec{0} & \frac{\vec{w}_s}{x_{st}}\\
\vec{0} & 0 & -\frac{\vec{w}_s^T}{x_{st}} & 0\tikzmark{right2} \\
\end{bmatrix}
\begin{bmatrix}
\vec{y}_{m} \\
y_{mt} \\
\vec{y}_{s} \\
y_{st} 
\end{bmatrix}
\DrawBox[thick, red ]{left1}{right1}{\textcolor{red}{\footnotesize$\vec{D}_m$}}
\DrawBoxx[thick, red]{left2}{right2}{\textcolor{red}{\footnotesize$\vec{D}_s$}},\\
\label{eq:power_preserving_interconnection}
\end{equation} \\
where $\vec{w}_m \in \mathbb{R}^{n_{y_m}}$ and $\vec{w}_s \in \mathbb{R}^{n_{y_s}}$ are steer values computed by the master and slave controllers, based on the states of the two tanks ${x_{mt} \text{ and } x_{st}}$. The matrices $\vec{D}_m$ and $\vec{D}_s$ that interconnect the tanks to the respective systems, are in the form of \cref{power preserving interconnection prototype}.
The resulting bilateral teleoperator energy rate matches the power flowing from master and slave devices into the master and slave controllers, indeed
\begin{equation}
\dot{\mathcal{H}}_{mt} + \dot{\mathcal{H}}_{c} + \dot{\mathcal{H}}_{st} = -\vec{u}_{mc}^T\vec{y}_{m} -\vec{u}_{sc}^T\vec{y}_{s}.
\end{equation}
where $\dot{\mathcal{H}}_c = P_{mt}^- - P_{mt}^+ + P_{st}^- - P_{st}^+$ is the energy stored in the communication channel. Therefore, the bilateral teleoperator is passive with respect to input $(-\vec{u}_{mc}, -\vec{u}_{sc})$ and output $(\vec{y}_m, \vec{y}_s)$. 
However, if the energy tanks are depleted, there is no possibility to passively implement the desired action. Therefore, the only requirement that we need to impose to preserve the passivity of the system is that the energy level inside each tank is positive. 
In the next section, we will show how we impose this requirement in the master and slave controllers.

\section{HYBRID TELEOPERATION CONTROL}
\label{sec:hybrid_teleoperation_control}

In the following sections we provide the implementation details of the proposed setup. First, we introduce the model of the slave, which is described by the equations of motion of a multi-limbed floating-base system:
\begin{equation}
\begin{split}
\vec{M}_u(\vec{q})\dot{\vec{v}} + \vec{b}_u(\vec{q},\vec{v}) &= \vec{J}_{c_u}^T(\vec{q})\vec{\lambda}\\
\vec{M}_a(\vec{q})\dot{\vec{v}} + \vec{b}_a(\vec{q},\vec{v}) &=  \vec{J}_{c_a}^T(\vec{q})\vec{\lambda} + \vec{u}_{sc},
\end{split}
\label{eq: dynamic model slave}
\end{equation}
where $\vec{q} \in SE(3) \times \mathbb{R}^{n_a}$ and $\vec{v} \in \mathbb{
R}^{6+n_a}$ are the robot generalized coordinates and velocities, $\vec{M}$ is the mass matrix, $\vec{b}$ includes the  nonlinear  effects (i.e. Coriolis, centrifugal, and gravitational terms), and $\vec{u}_{sc}$ is the  vector of  actuation  torques. Additionally, $\vec{J}_c$ is  a matrix  of  stacked contact  Jacobians,  while $\vec{\lambda}$ is the vector of contact wrenches. The subscripts $u$ and $a$ correspond to the unactuated and actuated parts of the defined quantities, respectively. 
We formulate the slave whole-body control problem as in \cite{bellicoso2016perception}, where a hierarchical QP is solved to determine the vector of desired generalized accelerations ($\dot{\vec{v}}^*$), and desired contact forces $(\vec{\lambda}^*)$. We denote the vector of optimization variables with ${\vec{\xi} = ({\dot{\vec{v}}^*}, {\vec{\lambda}^*})}$. A set of equality tasks encode dynamic consistency, no slipping or separation of stance legs, as well as tracking of base and limb motion references. Additionally, a set of inequalities is defined to encode friction cone constraints, joint torque limits, in addition to the passivity conditions. The priorities of each task are reported in Table~\ref{tab:table_priority_wbc} and will be discussed later in this section.
It is worth mentioning that the reference commands from the master device are not directly sent to the whole-body controller, but are given as inputs to an intermediate motion planner \cite{sleiman2020MPCplanner}. Briefly, this planner solves a model predictive control (MPC) problem to generate optimal motion references for the robot's base and limbs. These references are then tracked by the passivity-preserving QP hierarchical controller. Given the optimal solution $\vec{\xi}$ of the WBC, the joint torques are computed by inverting the desired dynamics:
\begin{equation}
\vec{u}_{sc} = [\vec{M}_a(\vec{q}) \hspace{0.2cm} - \vec{J}_{c_a}^T(\vec{q})]\cdot\vec{\xi} + \vec{b}_a(\vec{q}, \vec{v}).
\label{inverse dynamics control law}
\end{equation}
\subsection{End-effector position control}
\label{subsec:end_effector_position_control}
\subsubsection{Master controller} 
Since the controller is implemented on an embedded unit, a discrete-time rule needs to be computed to update the energy tank level \cite{stramigioli2005sampled}:
\begin{align}
&\mathcal{H}_{mt}(k) = \int_{0}^{k\Delta T}  \left(u_{mt} + \frac{P_{mt}^{+} - P_{mt}^{-}}{x_{mt}}\right)y_{mt} dt \notag \\ &= 
\int_{0}^{k\Delta T} -\vec{w}_m^T(t)\vec{y}_m(t) dt + \int_{0}^{k\Delta T} [P^+_{mt}(t) - P^-_{mt}(t)] dt \notag  \\ &= \sum_{j=1}^{k} -\vec{w}_m^T(j-1)\Delta \vec{x}_{m}(j) + \Delta\mathcal{H}^+_{mt}(j) - \Delta\mathcal{H}^-_{mt}(j) \notag  \\& = 
\mathcal{H}_{mt}(k-1) - \vec{w}_{m}^T(k-1)\Delta \vec{x}_m(k) + \Delta\mathcal{H}_{mt}(k), 
\end{align}
with ${\Delta \vec{x}_m(k) = \vec{x}_m(k) - \vec{x}_m(k-1)}$, ${\vec{y}_m = \dot{\vec{x}}_m}$, and ${\Delta\mathcal{H}_{mt}(k)}$ is the overall exchanged energy at time $k$.
Let $\vec{F}_c$ be the desired haptic feedback, which is updated based on feedback information from the slave's end-effector. 
We compute the master-side steer value $\vec{w}_m$ as a function of both the state of the tank and the desired haptic feedback:
\begin{equation}
 \vec{w}_m(k) =
\begin{cases}
\vec{F}_c(k) & \text{if } \mathcal{H}_{mt}(k) \ge \zeta\\
\frac{x_{mt}^2(k)}{\gamma^2}\vec{F}_c(k) -\vec{\Gamma}\cdot {\dot{\vec{x}}}_m & \text{if } 0 < \mathcal{H}_{mt}(k) < \zeta
\end{cases}
\label{eq:position_mode_master_control_input}
\end{equation}
where $\zeta > 0$ is the minimum energy level in the tank introduced to avoid singularities in the solution, ${\gamma = \sqrt{2\cdot\zeta}}$, and $\vec{\Gamma}\succ 0$ is a diagonal matrix whose entries are equal to ${\alpha\cdot(\zeta - \mathcal{H}_{mt}(k))}$ with $\alpha > 0$. When the value of the tank decreases below this threshold, $\vec{F}_c$ is modulated according to the state of the tank and, additionally, a dissipative term is added to extract energy from the master. The type of modulation proposed here offers the advantage of being smooth, and does not strictly require any estimate of the velocity at the next sample. Generally, up until the next sample is available, there is no way to act on a loss of passivity. However, if $\zeta$ is chosen as the maximum allowable energy to spend, this problem is avoided.
\begin{table}[t]
  \begin{center}
     \caption{List of tasks used in the WBC. Each task is assigned a priority. The tasks introduced in this work are highlighted, with reference to the corresponding equations.}
    \label{tab:table_priority_wbc}
    \begin{tabular}{ c l }
      \hline 
      Priority & Task\\
      \hline 
      1 & Equations of motion\\
        & Torque limits\\
        & Friction cone\\
        & Zero acceleration at the contact\\
     2 & \textcolor{blue}{Base motion direction} \eqref{eq: base direction preservation}\\
     & \textcolor{blue}{Arm motion direction} \eqref{eq: direction acceleration arm}\\
     3 & \textcolor{blue}{Base passivity constraint} \eqref{eq: passivity robot in velocity mode}\\
      & \textcolor{blue}{Arm passivity constraint} \eqref{eq: passivity constraint}\\
     & Base motion tracking\\
     & Limbs motion tracking\\
     \hline 
    \end{tabular}
  \end{center}
  \label{table 2}
\end{table}
\subsubsection{Slave controller}
At the slave side, a similar update for the tank can be obtained as follows:
\begin{equation}
\begin{split}
&\mathcal{H}_{st}(k) = 
\mathcal{H}_{st}(k-1)-{\vec{w}_s}^T(k-1) \Delta \vec{q}_s(k) + \Delta\mathcal{H}_{st}(k).
\end{split}
\label{slave_tank_update_equation}
\end{equation}
At each sample, $\mathcal{H}_{st}$ is updated with the previously applied torques $\vec{w}_s(k-1) = \vec{u}_{sc}(k-1)$, the joint displacement $\Delta \vec{q}_s(k) = \vec{q}(k) - \vec{q}(k-1)$, and the exchange with the master tank $\Delta \mathcal{H}_{st}(k) = \Delta\mathcal{H}_{st}^+(k) -\Delta\mathcal{H}_{st}^-(k)$.
The control law \eqref{inverse dynamics control law} needs to impose that the energy level inside the tank at each instant of time is larger than zero. Since the energy exchange with the master tank $\Delta \mathcal{H}_{st}$  at sample $k+1$ is unknown, we impose the following inequality constraint:
\begin{equation}
\mathcal{\hat{H}}_{st}(k+1) = \mathcal{H}_{st}(k) - \vec{u}_{sc}^T(k) \dot{\vec{q}}(k)\Delta T  \mbox{ } \geq \epsilon,   
\label{passivity condition at slave side in position mode}
\end{equation}
where $\mathcal{\hat{H}}_{st}$ is an estimate of the tank level at time $k+1$, $\epsilon > 0$ is a threshold for the minimum possible energy level, and the joint displacements $\Delta \vec{q}_s(k+1)$ are estimated from the current joint velocities. 

During teleoperation of the arm end-effector, we tolerate small variations of the torso, resulting in \eqref{passivity condition at slave side in position mode} being mainly affected by the joints of the arm. Under this consideration, \cref{passivity condition at slave side in position mode} can be rewritten taking \eqref{inverse dynamics control law} into account, as:
\begin{equation}
\small
  	\begin{bmatrix}\dot{\vec{q}}_r^T\vec{M}_{r} & -\dot{\vec{q}}_r^T\vec{J}^T_{ {c_r}}\end{bmatrix}\vec{\xi} \le  \frac{\mathcal{H}_{st}(k)-\epsilon}{\Delta T}- \dot{\vec{q}}_r^T(\vec{b}_{r}-\vec{g}_r)
  	\label{eq: passivity constraint}
\end{equation}
where the subscript $r$ corresponds to the arm degrees of freedom, $\vec{M}_r$ includes the last $n_r$ rows of $\vec{M}_a$, and $\vec{g}_{r}$ are the arm gravitational torques. These can be subtracted from the energy budget since such a compensation preserves passivity.

Let $\vec{q}^*_r$ and $\dot{\vec{q}}^*_r$ be the arm joint references computed by the MPC planner. A desired acceleration is generated with a PD control law:
\begin{equation}
\ddot{\vec{q}}_r^{des} = \vec{K}_p(\vec{q}_r^* - \vec{q}_r) +\vec{K}_d(\dot{\vec{q}}_r^* -\dot{\vec{q}}_r).
\label{eq: desired joint acceleration}
\end{equation}
An equality task is then formulated to track the resulting desired acceleration
\begin{equation}
[\vec{O}_{n_{r}\times n_z} \mbox{ }\mbox{ }\vec{\mathbb{I}}_{n_r\times n_r} \mbox{ }\mbox{ }\vec{O}_{n_r \times n_\lambda}]\cdot \vec{\xi} = \ddot{\vec{q}}_r^{des},
\label{eq: tracking task arm}
\end{equation}
where $n_z = n_a + 6 - n_r$.
To avoid having the passivity task invert the direction of the desired accelerations, we impose an additional constraint on the arm's motion, given by:
\begin{equation}
[\vec{O}_{n_r\times n_z} \mbox{ }\mbox{ }\vec{Q}_r \mbox{ }\mbox{ }\vec{O}_{n_r\times n_{\lambda}}] \cdot \vec{\xi} \ge \vec{0} ,   
\label{eq: direction acceleration arm}
\end{equation}
where $\vec{Q}_r \in \mathbb{R}^{n_r\times n_r}$ is a diagonal matrix with diagonal entries equal to the desired joint accelerations as in \eqref{eq: desired joint acceleration}.
A higher priority (see Table~\ref{tab:table_priority_wbc}) is assigned to \eqref{eq: direction acceleration arm} when compared to the passivity and tracking tasks. In this way we ensure that the desired joint-space motion direction is preserved, while jointly optimizing for passivity and transparency (i.e tracking) with a lower priority. One could expect that the passivity task should be at the highest priority. However, as pointed out in \cite{selvaggio2019passive}, since passivity is only sufficient for stability, strictly enforcing passivity could lead to an overly conservative behavior. Hence, for our control architecture, enforcing stability at a lower priority allows to respect the stability requirements, while still guaranteeing a good tracking performance.

\subsection{Base velocity control}
\label{subs: controllers_velocity_mode}
\subsubsection{Master controller}
The base velocity teleoperation problem is challenging since the haptic device has a limited workspace. Ideally, it would be desirable to map master positions to slave velocity commands. However, the master device is passive with respect to the power port $(\vec{u}_{mc}, \dot{\vec{x}}_m)$ rather then $(\vec{u}_{mc}, \vec{x}_m)$. As in \cite{lee2011feedback}, we overcome this issue by rendering the master passive with respect to the port $(\vec{u}_{mc}, \vec{r}_m)$, where the output $\vec{r}_m$ is defined as: ${\vec{r}_m = \dot{\vec{x}}_m + \vec{\Lambda}\vec{x}_m}$, with ${\vec{\Lambda} \succ 0}$; and a local control action is added to $\vec{u}_{mc}$ as $-\vec{B}\dot{\vec{x}}_m-\vec{K}\vec{x}_m$. The matrix $\vec{B}$ is chosen such that $\vec{B} \succ \vec{M}_m\vec{\Lambda}$, with $\vec{M}_m$ being the mass matrix of the haptic device, $\vec{K}\succ 0$. As shown in \cite{franchi2012bilateral}, the signal $\vec{r}_m$ can be made proportional only to the scaled position if the contribution of velocity is negligible due to the choice of $\vec{\Lambda}$.

\subsubsection{Slave controller}
The slave degrees of freedom controlled in velocity mode are the lateral and longitudinal velocities of the base. No references are sent along the vertical direction, however we still allow for feedback. In this way, if the base of the robot is disturbed along this direction, the operator could still perceive the applied force. Since the floating-base of a legged robot is driven by generating contact forces on the environment, we impose the energetic limitation with respect to the $xy$ (non-vertical) forces exerted by the stance legs: 
\begin{equation}
\mathcal{\hat{H}}_{st}(k+1) = \mathcal{H}_{st}(k) - (\vec{J}^T_{{u}_{xy}}\vec{\lambda})^T\dot{\vec{x}}_{s_{xy}}\Delta T \geq \epsilon,   
\label{passivity condition at slave side in velocity mode}
\end{equation}
where $\vec{J}^T_{u_{xy}}$ includes the first two rows of ($\vec{J}_{c_u}^T$), and $\dot{\vec{x}}_s$ is the base velocity. Using \eqref{eq: dynamic model slave}, we can rewrite \eqref{passivity condition at slave side in velocity mode} as:
\begin{equation}
\small
    [\vec{\dot{x}}_{s_{xy}}^T\vec{M}_{xy} \;\;\; \vec{0}_{2 \times n_{\lambda}}]\cdot\vec{\xi} \leq \frac{\mathcal{H}_{st}(k)-\epsilon}{\Delta T} \;-\dot {\vec{x}}_{s_{xy}}^T(\vec{b}_{xy}-\vec{g}_{xy}),
    \label{eq: passivity robot in velocity mode}
\end{equation}
which can be added as a task in the hierarchical QP.
As in Sec.~\ref{subsec:end_effector_position_control}, the desired linear accelerations calculated by the motion planner should be preserved. These accelerations are expressed in a specific frame, named \textit{control frame} $C$ \cite{bellicoso2016perception}, which changes according to the slope of the terrain and the base orientation. To prevent changes in desired references, we add the following additional inequality constraint to the stack of tasks:
\begin{equation}
\begin{bmatrix}\vec{Q}_{xy} \vec{J}_{u_{xy}} & \vec{0}_{2\times n_{\lambda}}
\end{bmatrix} \cdot \vec{\vec{\xi}} + \vec{Q}_{xy} \dot{\vec{J}}_{u_{xy}}\vec{v} \ge \vec{0},
\label{eq: base direction preservation}
\end{equation}
where $\vec{Q}_{xy} \in \mathbb{R}^{2\times2}$ is a diagonal matrix with entries equal to the elements of the desired linear acceleration ${}_{C}\Ddot{\vec{r}}^{des}_{xy}$, which is computed as follows:
\begin{equation}
{}_{C}\Ddot{\vec{r}}^{des}_{xy} = {}_{C}\Ddot{\vec{r}}_{xy}^* + \vec{K}_d({}_{C}\dot{\vec{r}}_{xy}^* -{}_{C}\dot{\vec{r}}_{xy}) + \vec{K}_p({}_{C}{\vec{r}}_{xy}^* - {}_{C}{\vec{r}}_{xy}).\\
\end{equation}


\section{EXPERIMENTAL RESULTS}
\label{sec:experimental_results}
We validated our proposed teleoperation framework through experiments performed on a real robot platform with an artificial delay introduced in the network. Videos of all the experiments are available in the attached multimedia material. The master device used for the experiments is the Force Dimension Omega.6 haptic device\footnote{\href{https://www.forcedimension.com/products/omega}{https://www.forcedimension.com/products/omega}}. The slave robot is ANYmal -- a torque-controlled quadrupedal robot -- equipped with a 4 DOF robotic arm. The control loop runs at 1 kHz on the haptic device, while the ANYmal whole-body controller and the state estimator run at 400 Hz on the robot's onboard  computer  (Intel  Corei7-8850H  CPU@4GHz  hexacore  processor). 
Each experiment is performed without and with the proposed passivity framework.
We provide results for end-effector position control and base velocity control, with an end-to-end added delay of 60 ms for the two scenarios. This delay was observed as the average communication delay between the University of Twente and our laboratory in Zurich.
\subsection{End-effector position control}
\label{subsec:experiments_in_position_mode}
In this experiment, the human operator moves the haptic device along a predefined trajectory, defined in such a way to span the whole workspace up to the boundaries of the haptic device.
\begin{figure}[t]
\centering
\includegraphics[width=\columnwidth]{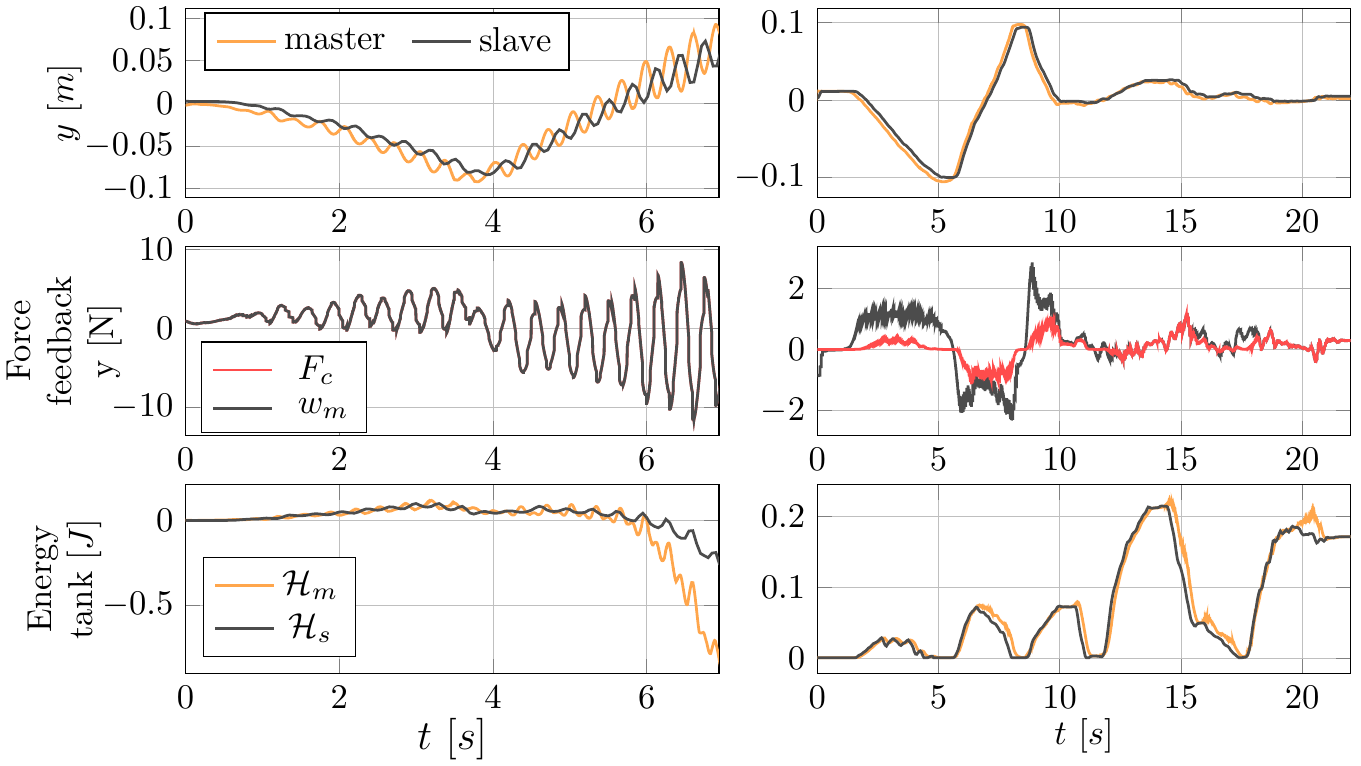}
\caption{Master and slave end-effector position, force and energy tanks levels with passivity layer \textit{(right)}, and without \textit{(left)}, during an experiment of end-effector control.}
\label{fig: devices in position control}
\vspace{-0.2cm}
\end{figure}
For this experiment, the local force feedback used in Eq.~\eqref{eq:position_mode_master_control_input} was generated from the displacement error between the master and the slave measured position: $\vec{F}_c(k) = \vec{K}_p(\vec{x}_s(k-k_d) - \vec{x}_m(k))$
where $k_d$ is the time delay. 
Results for this experiment are displayed in \cref{fig: devices in position control} where, for the sake of conciseness, only the plots along the y direction are reported.
It can be seen that, without the passivity layer, the two devices show oscillatory signals that are out-of-phase. In this condition, to avoid damages to the hardware, the system is frozen as soon as the oscillations start to grow. 
Another indicator of unstable behavior is observed in the plot of the energy tank level, where a drop to a negative value occurs. Conversely, with passivity layer, this behavior is avoided and leads to smoother master and slave trajectories.
\subsection{Base velocity control}
\label{subsec:experiments_in_velocity_mode}
In this experiment, we control the base velocity of the slave. To avoid hardware damage we limit the maximum commanded velocity to 0.2 m/s. For this test, the local force feedback used in \cref{eq:position_mode_master_control_input} was generated from the velocity tracking error at the slave side: ${\vec{F}_c(k) = \vec{K}_v(\vec{v}_m(k-k_d) - \vec{v}_s(k))}$,
where $\vec{v}_m$, and $\vec{v}_s$ denote the master velocity, and the slave base velocity, respectively. $\vec{v}_m$ is computed from the master position using the mapping from \cref{subs: controllers_velocity_mode}. This haptic feedback allows to realistically perceive external forces, and to capture their effect without the need for any haptic sensory feedback.
\begin{figure}[t]
\centering
\hspace*{-0.1\linewidth}
\begin{tikzpicture}
\node[inner sep=0pt] (plotTrajectoryPositionMode) at (-0.7,0)
    {\includegraphics[scale=0.7,width=\columnwidth]{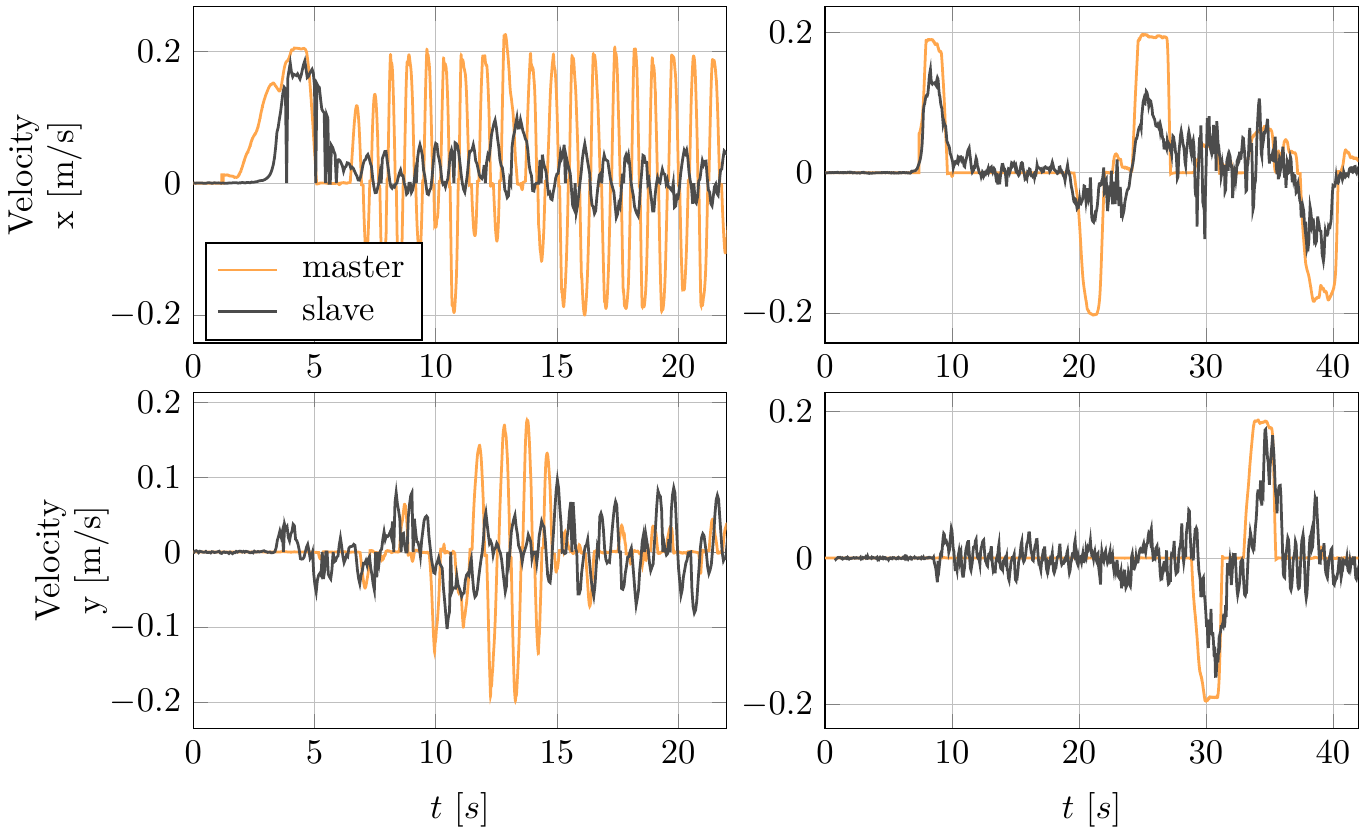}};
\end{tikzpicture}
\caption{Master and slave $x$-$y$ velocity profiles with passivity layer \textit{(right)}, and without \textit{(left)} in base velocity control.}
\label{fig: velocity profile x-y in rate mode control.}
\vspace{-0.2cm}
\end{figure}
\begin{figure}[t]
    \centering
    \hspace*{-0.1\linewidth}
\includegraphics[scale=0.8,width=\columnwidth]{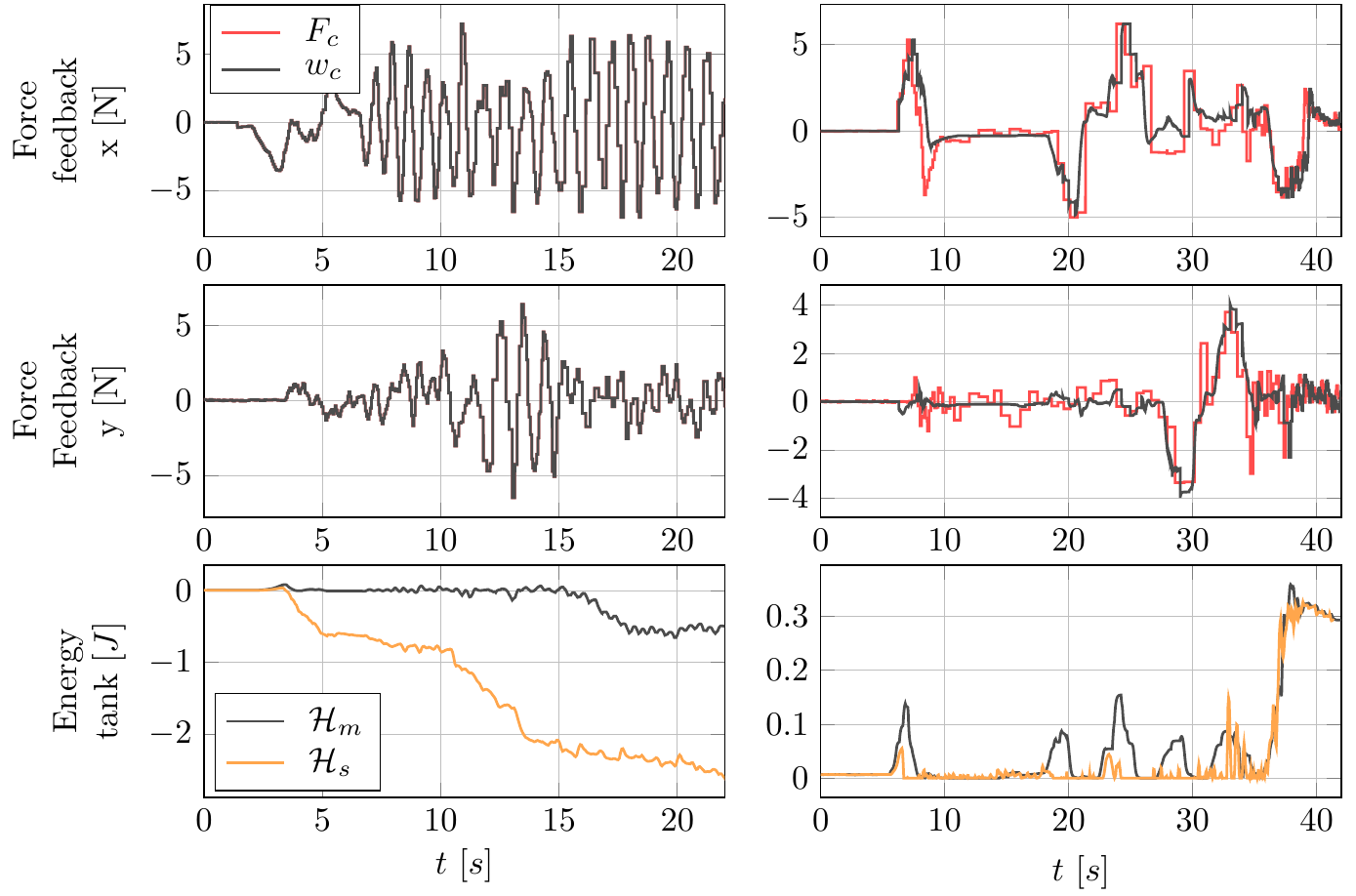}
\caption{Force and energy tanks levels with passivity layer \textit{(right)}, and without \textit{(left)} in base velocity control. }
\label{fig: energy profile in rate mode control.}
\vspace{-0.3cm}
\end{figure}
It can be noted from \cref{fig: velocity profile x-y in rate mode control.} that, with passivity layer, the velocity tracking error increases as the master velocity is about to reach $0.2 \ \text{m/s}$. While the master velocity is about reaching this peak, the mean absolute percentage tracking error in velocity is 6.98\% without passivity layer, and only 1.1\% with passivity layer. The former scenario happens in order to avoid violating the passivity condition as seen in the energy level plots of \cref{fig: energy profile in rate mode control.}, which in turn leads to the energetic penalization of the contact forces along the directions where they create motion. 
As shown in Fig.~\ref{fig: velocity profile x-y in rate mode control.}, if passivity is not enforced, the force feedback causes the haptic device to become unstable, and this in turn compromises tracking. Enforcing passivity allows to avoid this problem and results in a stable teleoperation system.
\begin{figure}[t]
\centering
\hfill
\includegraphics[width=4.1cm]{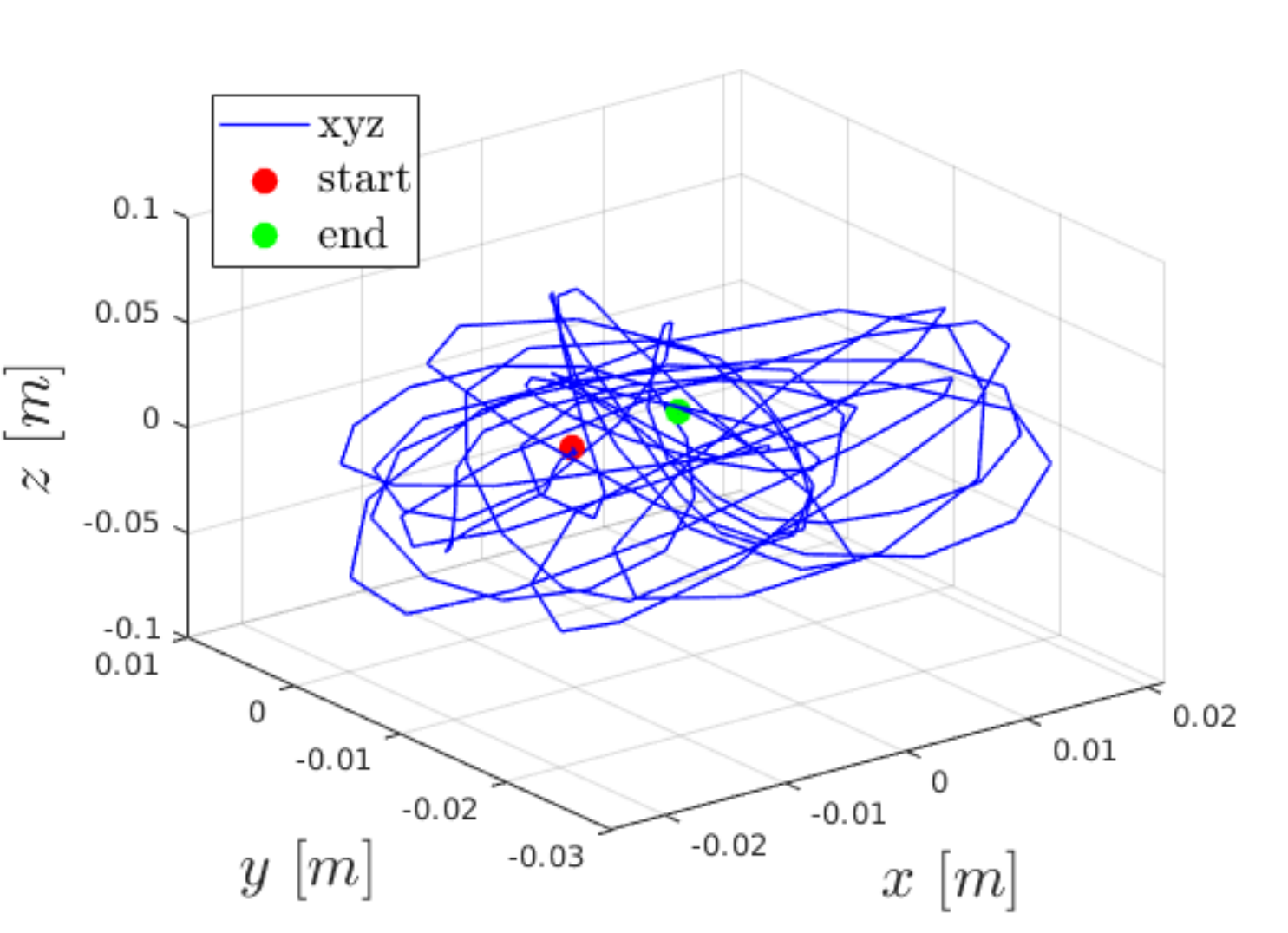}
\hfill
\includegraphics[width=4.1cm]{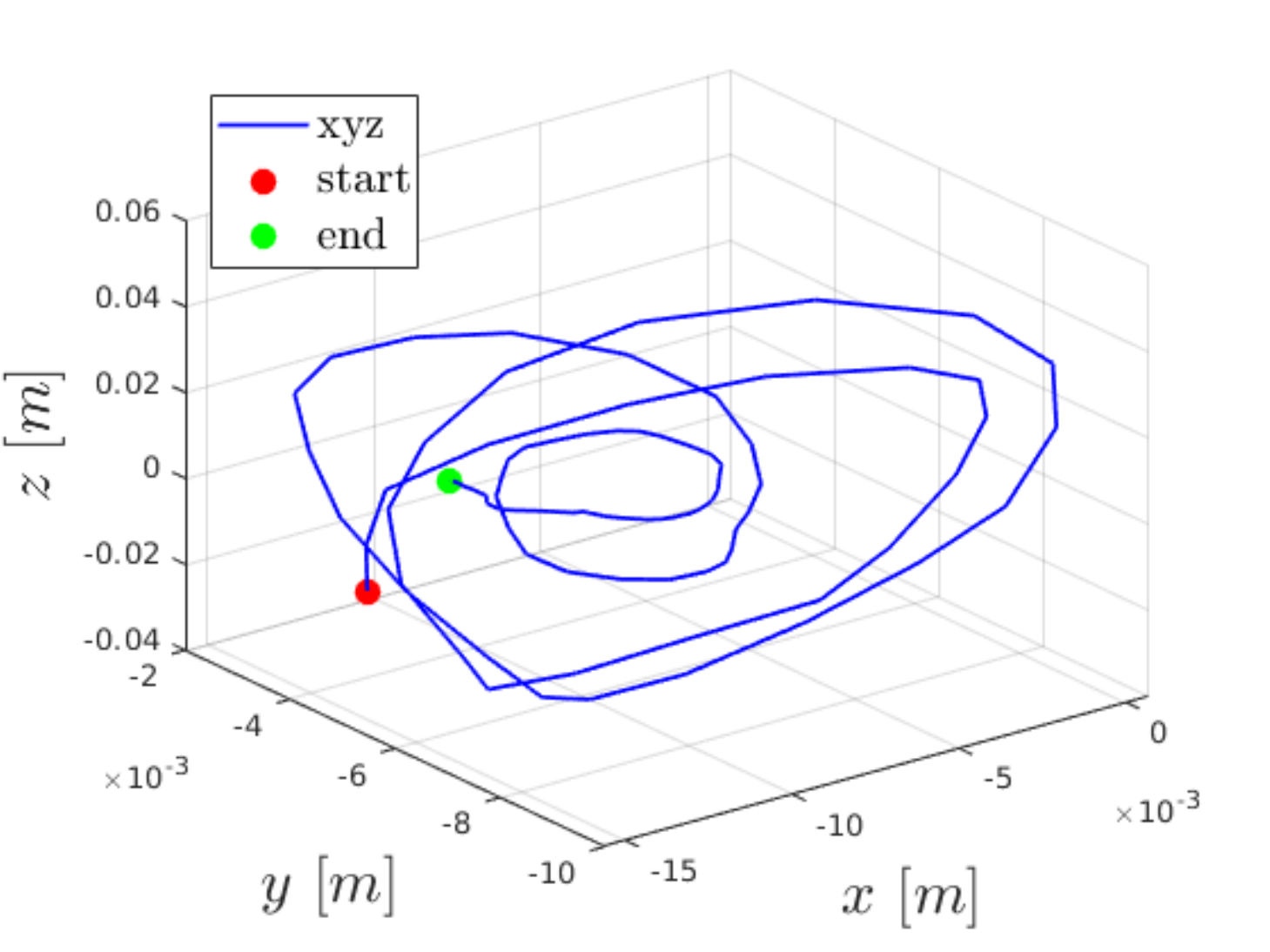}
\hfill
\caption{3D position trajectories for the slave end-effector during two experiments of haptic interaction with a human, with energy tanks off \textit{(left)}, and on \textit{(right)}.}
\label{fig: hand shaking}
\vspace{-0.7cm}
\end{figure}
\subsection{Haptic interaction during end-effector control}
\label{subsec:haptic_interaction_position_mode}
In this experiment, we perform a haptic interaction test with a human at the slave side shaking the robot's hand. Haptic feedback was generated as in Sec.~\ref{subsec:experiments_in_position_mode}. For the sake of conciseness, we only present results for the slave response (\cref{fig: hand shaking}). From both the plots and the video attachment, it can be verified that, with passivity, the trajectories are smoother and the system remains stable. Gains for the master controller were tuned in such a way that enables the operator to perceive low magnitude forces acting on the slave end-effector.
We realized that, if it is desired that the operator feels the hand-shake, these gains tend to be high, leading to a more unstable behavior. Indeed, in haptic teleoperation, instabilities are due not only to time delays, but also to other sources, such as those deriving from a controller tuning that aims to enhance transparency. However, it can be noted from~\cref{fig: hand shaking} that our formulation is also effective in handling this additional source of instability.


\section{CONCLUSIONS}
\label{sec:conclusions}
In this paper, we presented an approach to ensure the stability of a haptic teleoperation system for a quadrupedal mobile manipulator. The crucial point to ensure such a property is to manage a potential active behavior of the bilateral teleoperator. We addressed such a problem for the bilateral teleoperation of a general floating-base system, in the presence of destabilizing time delays. To this end, energy tanks were used as energy observers and included in the formulation of a passive QP controller at the slave side. 
We experimentally demonstrated that, in the presence of time delays, the proposed passivity formulation based on energy tanks effectively leads to more stable operations. 
One interesting direction for future work includes adding feedback from force sensor measurements, and extending the proposed method to perform tele-manipulation tasks, such as remotely turning valves or opening doors.

\bibliographystyle{./bibtex/IEEEtran} 
\bibliography{root}
\addtolength{\textheight}{-12cm}   
\end{document}